\documentclass[10pt,journal,compsoc]{IEEEtran}

\ifCLASSOPTIONcompsoc
  \usepackage[nocompress]{cite}
\else
  \usepackage{cite}
\fi

\ifCLASSINFOpdf
  \usepackage[pdftex]{graphicx}
\else
  \usepackage[dvips]{graphicx}
\fi

\usepackage{amsmath}

\usepackage{array}
\usepackage{mdwmath}
\usepackage{mdwtab}
\usepackage{eqparbox}
\usepackage{booktabs}
\usepackage{multirow}
\usepackage{tabularx}
\usepackage{tabulary}
\usepackage{graphicx}
\usepackage{wrapfig}
\usepackage{longtable}
\usepackage{makecell} 

\ifCLASSOPTIONcompsoc
  \usepackage[caption=false,font=footnotesize,labelfont=sf,textfont=sf]{subfig}
\else
  \usepackage[caption=false,font=footnotesize]{subfig}
\fi

\usepackage{siunitx}
\usepackage{fixltx2e}
\usepackage{dblfloatfix}
\ifCLASSOPTIONcaptionsoff
  \usepackage[nomarkers]{endfloat}
 \let\MYoriglatexcaption\caption
 \renewcommand{\caption}[2][\relax]{\MYoriglatexcaption[#2]{#2}}
\fi

\usepackage{url}

\newcommand\MYhyperrefoptions{bookmarks=true,bookmarksnumbered=true,
pdfpagemode={UseOutlines},plainpages=false,pdfpagelabels=true,
colorlinks=true,linkcolor={black},citecolor={black},urlcolor={black},
pdftitle={Bare Demo of IEEEtran.cls for Computer Society Journals},
pdfsubject={Typesetting},
pdfauthor={Michael D. Shell},
pdfkeywords={Computer Society, IEEEtran, journal, LaTeX, paper,
             template}}
\ifCLASSINFOpdf
\usepackage[\MYhyperrefoptions,pdftex]{hyperref}
\else
\usepackage[\MYhyperrefoptions,breaklinks=true,dvips]{hyperref}
\usepackage{breakurl}
\fi

\begin{document}

\title{BiofilmScanner: A Computational Intelligence Approach to Obtain Bacterial Cell Morphological Attributes from Biofilm Image}

\author{Md~Hafizur~Rahman,
        Md~Ali~Azam, Md~Abir~Hossen, Shankarachary Ragi, and~Venkataramana~Gadhamshetty
\IEEEcompsocitemizethanks{\IEEEcompsocthanksitem M.H. Rahman is with GM Core System Tools \& Automation, 29755 Louis Chevrolet Road, Warren, MI 48093\protect\\
E-mail: hafizur.raj@gmail.com
\IEEEcompsocthanksitem  M.A. Azam with GE HealthCare, 9900 W Innovation Dr, Wauwatosa, WI 53226, USA.
\IEEEcompsocthanksitem  M.A. Hossen is with the Department of Computer Science, University of South Carolina, Columbia, SC 29208, USA.
\IEEEcompsocthanksitem  S. Ragi was with the Department of Electrical Engineering, South Dakota Mines, Rapid City, SD 57701, USA.
\IEEEcompsocthanksitem  V. Gadhamshetty is with the Department of Civil and Environmental Engineering, South Dakota Mines, Rapid City, SD 57701, USA.}
}

\markboth{IEEE TRANSACTIONS ON COMPUTATIONAL BIOLOGY AND BIOINFORMATICS }%
{MH Rahman \MakeLowercase{\textit{et al.}}:BiofilmScanner: A Computational Intelligence
Approach to Obtain Bacterial Cell Morphological Attributes from Biofilm Image}


\IEEEtitleabstractindextext{%
\begin{abstract}
Desulfovibrio alaskensis G20 (DA-G20) is utilized as a model for sulfate-reducing bacteria (SRB) that are associated with corrosion issues caused by microorganisms. SRB-based biofilms are thought to be responsible for the billion-dollar-per-year bio-corrosion of metal infrastructure. Understanding the extraction of the bacterial cells' shape and size properties in the SRB-biofilm at different growth stages will assist with the design of anti-corrosion techniques. However, numerous issues affect current approaches, including time-consuming geometric property extraction, low efficiency, and high error rates. This paper proposes BiofilScanner, a Yolact-based deep learning method integrated with invariant moments to address these problems. Our approach efficiently detects and segments bacterial cells in an SRB image while simultaneously invariant moments measure the geometric characteristics of the segmented cells with low errors. The numerical experiments of the proposed method demonstrate that the BiofilmScanner is 2.1x and 6.8x faster than our earlier Mask-RCNN and DLv3+ methods for detecting, segmenting, and measuring the geometric properties of the cell. Furthermore, the BiofilmScanner achieved an F1-score of 85.28\% while Mask-RCNN and DLv3+ obtained F1-scores of 77.67\% and 75.18\%, respectively.
\end{abstract}

\begin{IEEEkeywords}
Deep learning, biofilm, image analysis, computer vision, sulfate-reducing bacteria.
\end{IEEEkeywords}}

\maketitle

\IEEEdisplaynontitleabstractindextext

%
\IEEEpeerreviewmaketitle

\ifCLASSOPTIONcompsoc
\IEEEraisesectionheading{\section{Introduction}\label{sec:introduction}}
\else
\section{Introduction}
\label{sec:introduction}
\fi

\begin{figure*}[!th]
\centering
\includegraphics[width=7in]{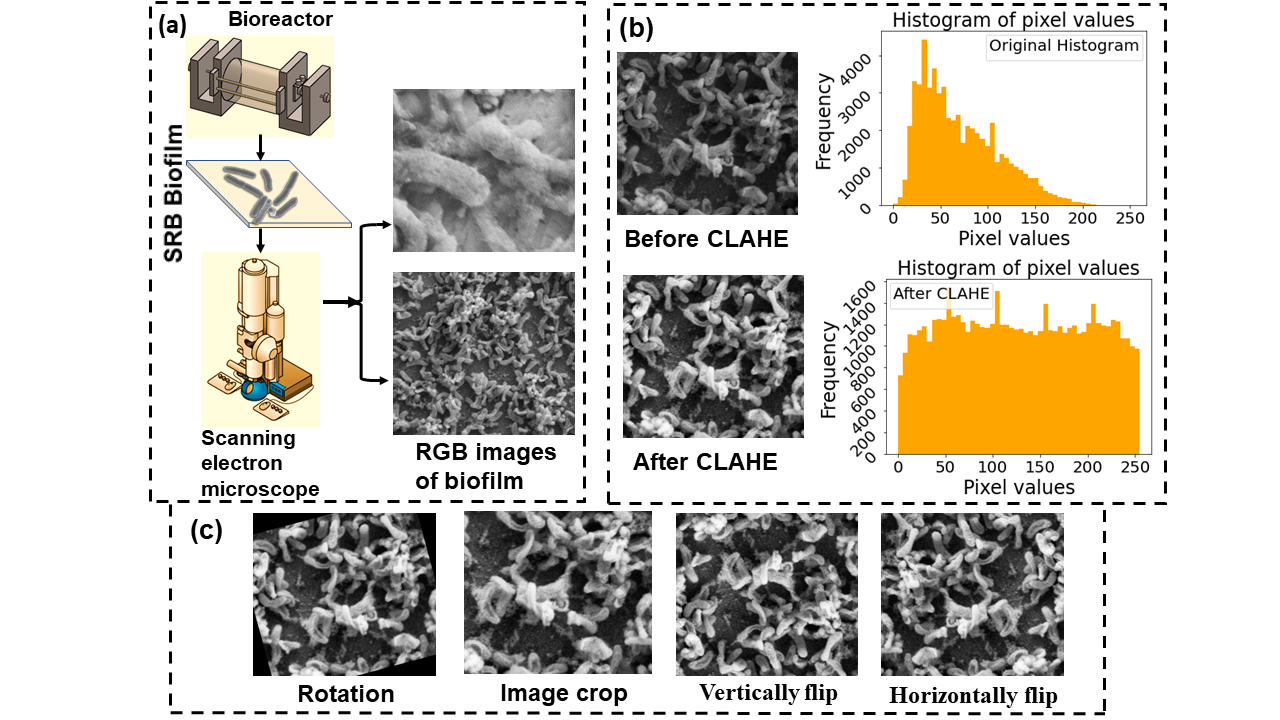}
\includegraphics[width=7in]{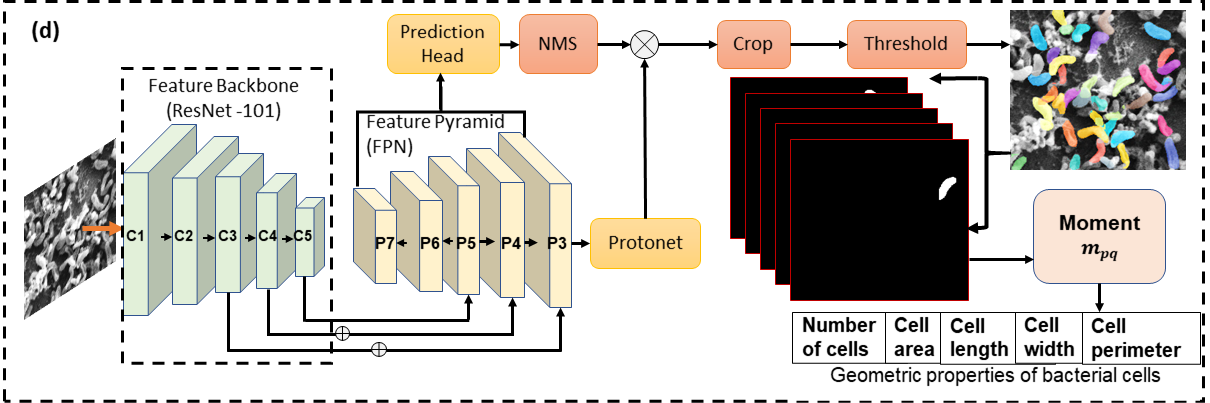}
\includegraphics[width=7in]{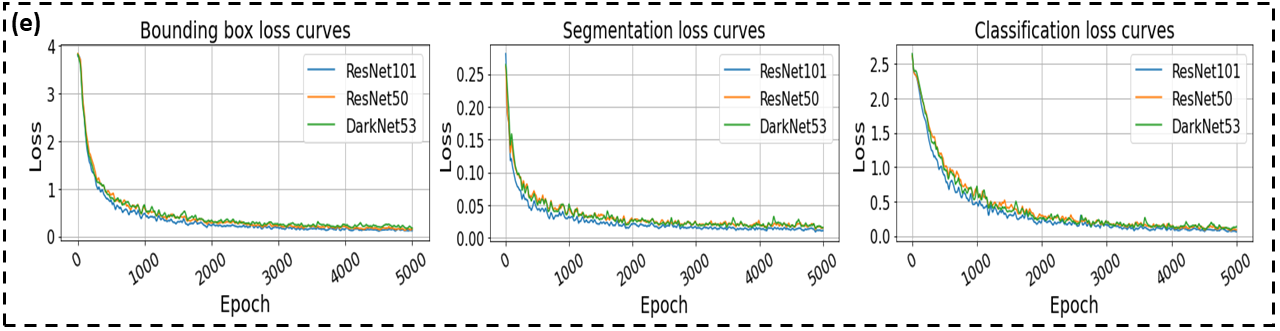}
\caption{(a) Dataset generation process at various growth stages of SRB biofilms with multiple scales; (b) Image pre-processing via CLAHE to increase the number of detecting bacterial cells; (c) Data augmentation to increase data volume and avoid overfitting; (d) Yolact-based deep learning model, for instance, segmentation of bacterial cells (e) learning curves for ResNet-101, ResNet-50, Darknet-53 (The presentation formation of the graphic was adapted from \cite{ragi2021artificial})}
\label{fig_1}
\end{figure*}

%
%

\IEEEPARstart{M}{icrobiologically} influenced corrosion (MIC) is a complex interplay of electrochemical, environmental, operational, and biological elements that frequently leads to significant material deterioration and corrosion in military applications, the marine sector, the oil industry, utilities, and transportation \cite{hou2017cost}\cite{li2015materials}\cite{lechevallier1990disinfecting}. Combining information from market evaluations, academic studies, and industry publications, it is estimated that biofilms generate over US\$4,000 billion a year, with around two-thirds of that amount expected to be damaged to corrosion worldwide in 2019 \cite{hofer2022cost}\cite{jacobs1998sulfide}\cite{camara2022economic}. MIC of metals is frequently attributed to sulfate-reducing bacteria (SRB) of the biofilm. Corrosion-resistant alloys, organic coatings, corrosion inhibitors, Q235 carbon steel produced through biomineralization, and anodic/cathodic protection are all used to prevent corrosion \cite{lou2021microbiologically}. Still, their drawbacks, such as high cost, significant contamination, and operational challenges, have not yet been fully overcome \cite{lou2021microbiologically}. Therefore, designing and developing further corrosion prevention strategies requires understanding the phenotypical growth characteristics of biofilms at different stages, specifically the size and geometry of the bacterial cells in biofilms on metal surfaces and how they adapt in hazardous situations associated with corrosion. Our objective in this study is to automatically extract the geometric characteristics (shape, number of cells, etc.) of SRB cells from images of the SRB-biofilm taken under a scanning electron microscope (SEM) at various growth phases. These geometric features are frequently extracted and measured using deep learning techniques or traditional image processing techniques, which are labor-intensive, prone to errors, and have low rates of efficiency. In order to efficiently automate this process \cite{hossen2021developing} \cite{hasan2020smart} \cite{chowdhury2022deepqgho}, our proposed algorithm makes use of computer vision and deep learning techniques to automatically extract each cell's geometric properties from images of biofilms.

\begin{figure*}[!h]
\centering
\includegraphics[width=7in]{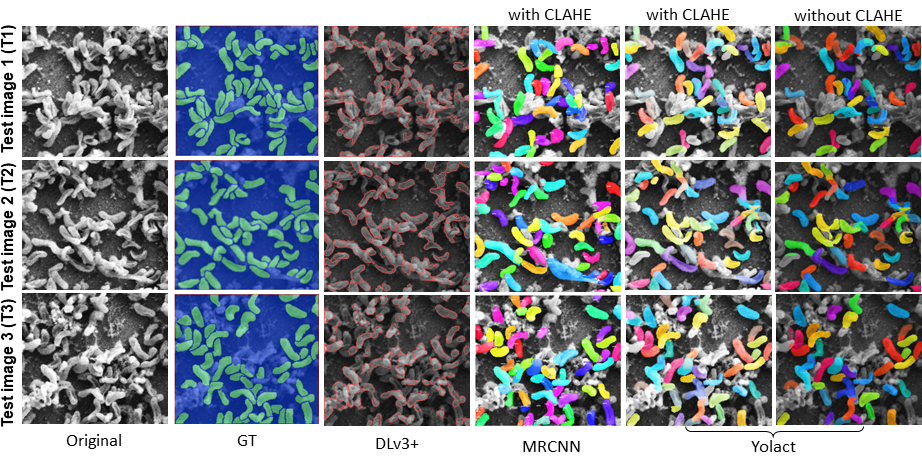}
\caption{From left to right column: raw SEM images of biofilm, ground truth for cell segmentation (manually labeled), automatic segmentation from DLv3+, instance segmentation via MRCNN, instance segmentation with image preprocessing via Yolact, and instance segmentation without image preprocessing via Yolact. ‘Test Image X’ refers to the SEM images generated in identical experimental conditions, where X is the image index. }
\label{img3}
\end{figure*}

\begin{figure*}[!h]
\centering
\includegraphics[width=7in]{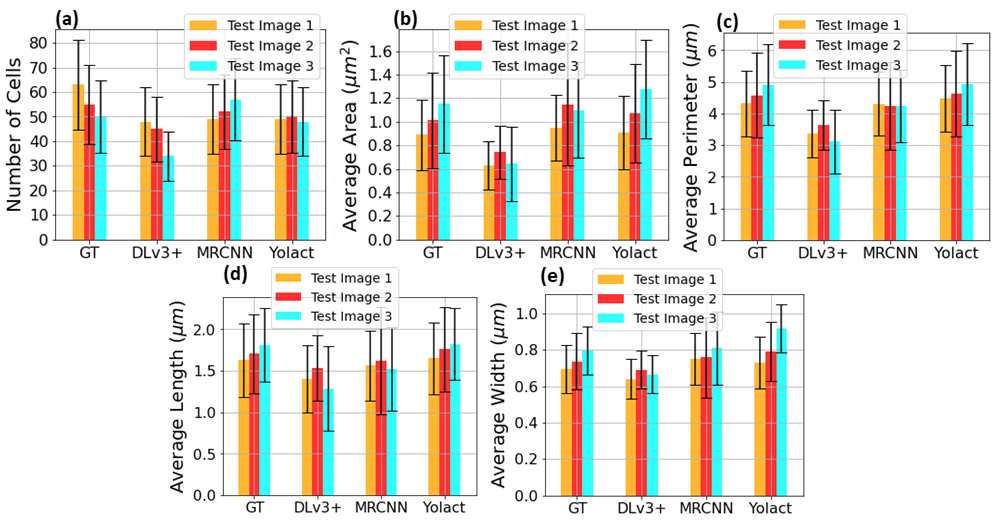}
\caption{Geometric properties of the segmented cells estimated via different segmentation approaches (color bars represent a row of Figure 4). (a) the number of cells; (b) the average area of the cells; (c) the average length of the cells; (d) the average width of the cells; (e) the average perimeter of the cells. The error bars represent the "one" standard deviation of the mean. Acronyms GT, DLv3+, MRCNN, and Yolact represent ground truth, DeepLabV3+, MRCNN, and Yolact respectively.}
\label{fig_4}
\end{figure*}

\begin{figure*}[!th]
\centering
\includegraphics[width=7in]{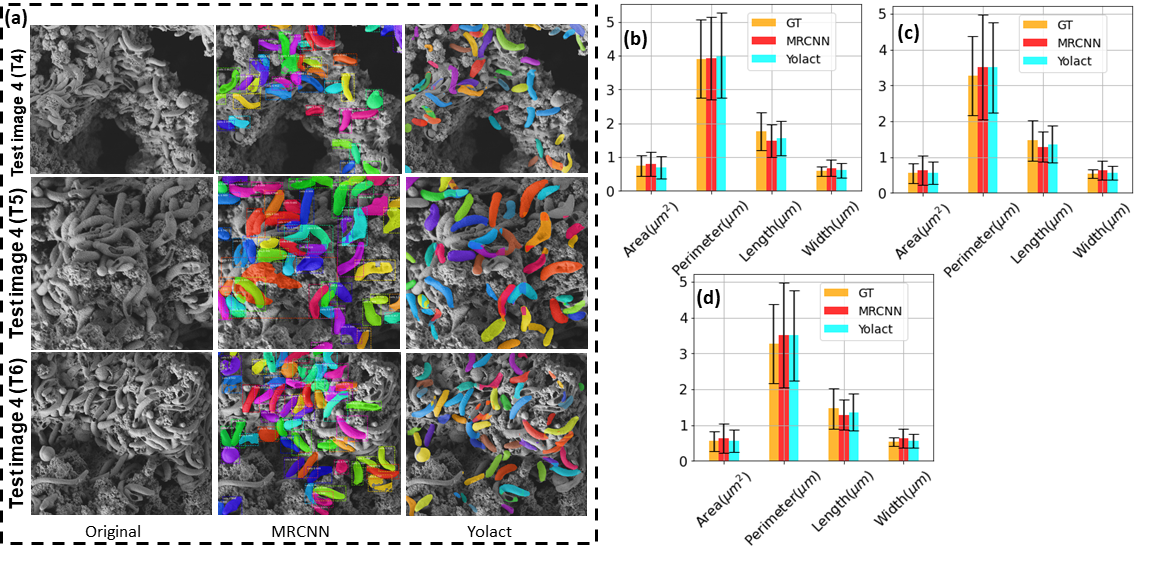}
\caption{Performance of Mask-RCNN, and Yolact for cell segmentation and cell size estimation on a biofilm system where images were captured on DA-G20 formed on 56.2\% cold-worked copper (a) From left to right: Raw SEM image of the biofilm after 70 days of growth; Instance segmentation of bacterial cells from MRCNN approach; Instance segmentation of bacterial cells from Yolact approach; (b) (c) (d) Estimation of size properties from - the moment invariants method applied on the segmented images from the ground truth, MRCNN, and Yolact (error bars represent standard deviation).}
\label{fig_5}
\end{figure*}

Microscopy image feature extraction and measuring the geometric properties of objects in the image have been successfully accomplished in the past using conventional image analysis tools for microscopy analysis, such as BiofilmQ \cite{hartmann2019biofilmq}, ImageJ \cite{prodanov2012automated}, BioFilm Analyzer \cite{bogachev2018fast}, Imaris \cite{Imaris}, etc. While the objects in microscope images are heterogeneous and have overlapping features, as they do in our case, traditional image analysis algorithms have a tendency to result in significant error rates. Additionally, the geometric computing steps of these methods are not entirely automated. As an alternative, deep learning approaches can be used to get beyond the shortcomings of conventional image processing methods~\cite{gopalakrishnan2023applications}. Deep learning-enhanced technologies offer a learning strategy that directly processes raw data and automatically learns representations while requiring fewer human interventions. The foundation of deep learning is a deep artificial neural network framework with multiple layers that can learn the high-dimensional hierarchical features of objects from training datasets using a backpropagation algorithm, typically used to train the network while minimizing the error between the predicted and actual labels \cite{paola1995review}. Due to these distinct and reliable characteristics, deep learning-assisted solutions are being developed in application fields such as medical image analysis \cite{hossen2021total}, speech recognition \cite{rabiner1993fundamentals}, self-driving cars \cite{dey2023intelligent} \cite{azam2021uav}, object detection \cite{zhao2019object}, semantic segmentation \cite{wang2018understanding}, instance segmentation \cite{correia2003objective} \cite{rahman2021deep}, and image generation \cite{zhao2019image}. Thus, deep learning methods such as MRCNN \cite{ragi2021artificial}, DLv3+ \cite{rahman2021deep}, and others \cite{sauls2019mother} have also been used for microscopy image analysis. These approaches, however, continue to be inefficient, have greater error rates, and are not time efficient.

The goal of this article is to develop an efficient deep learning-based model called BiofilmScanner to extract the geometric characteristics of the individual bacterial cells in an SEM biofilm image. We demonstrate that the microscale geometric properties of biofilms can be extracted with high accuracy using deep learning in combination with computer vision technologies. For challenges like bacterial cell segmentation, we use a deep learning technique called Yolact \cite{bolya2019yolact} (Fig. \ref{fig_1}). The moment invariants approach \cite{mukundan1998moment} is merged with the Yolact algorithm to extract the segmented cells' geometrical characteristics. Finally, we evaluate how well the BiofilmScanner tool performs in comparison to two commercial microscopy tools: MRCNN \cite{ragi2021artificial} and a DLv3+-based model \cite{rahman2021deep}.

\section{Method}
We adopt a State-of-the-Art deep learning model called Yolact \cite{bolya2019yolact} (based on instant segmentation) to segment bacterial cells in the SEM images. Yolact was then combined with the moment invariants method \cite{mukundan1998moment} to automatically extract the geometric size attributes of the segmented bacterial cells, such as area, length, width, and perimeter of the cells, as shown in Fig. \ref{fig_1}(d). Next, we compare the effectiveness of our approaches to our earlier research on MRCNN \cite{ragi2021artificial} and DLv3+ \cite{rahman2021deep}. At the Two-Dimensional Materials for Biofilm Engineering, Science, and Technology (2D-BEST) center, which is researching MIC prevention on technologically significant metals such as polymers \cite{krishnamurthy2015superiority}, polymer composites \cite{chilkoor2020maleic}, graphene \cite{chilkoor2020atomic}, and hexagonal boron nitride \cite{chilkoor2020hexagonal}, we compare the effectiveness of our approach to the ground-truth data, which is manually measured by subject-matter specialists. We assess the performance of our Yolact-based model \cite{bolya2019yolact} for cell segmentation, extraction of cell size, and execution time on a new microbial corrosion system where DA-G20 cells are grown on copper material in order to further verify its resilience and efficacy.

\subsection{Yolact}
YOLACT \cite{bolya2019yolact} (You Only Look At Coefficients) is a real-time instance segmentation algorithm that uses a single convolutional neural network (CNN) to predict both object bounding boxes and class-specific masks. The model is designed to be fast and efficient, making it suitable for use in real-time applications such as video surveillance, robotics, image analysis, and autonomous vehicles. YOLACT is a single-stage detector, which means that it does not require a separate region proposal network like other instance segmentation methods such as MRCNN. Instead, YOLACT uses a single CNN to predict both object locations and masks. The network is trained to predict a set of class-specific mask coefficients, which are then combined to generate the final masks. This approach reduces the computational complexity and improves the real-time performance of the model.

The structure of the YOLACT \cite{bolya2019yolact} network is shown in Fig. \ref{fig_1}(d) and the network consists of several key components. First, the feature extractor is a pre-trained CNN, such as ResNet-101 \cite{he2016deep}, that is used to extract features from the input image. The feature maps generated by the feature extractor are used as input to the rest of the network. Second, the base feature pyramid \cite{lin2017feature} is a set of feature maps that are shared by all object instances. These feature maps are used to predict the mask coefficients, which are then combined with the instance-specific feature maps to generate the final masks. Third, class-specific Mask Coefficients: the network is trained to predict a set of class-specific mask coefficients for each object instance. These coefficients are then used to generate the final masks. Fourth, instance-specific Feature Maps: the network also generates a set of instance-specific feature maps for each object instance. These feature maps are used in combination with the class-specific mask coefficients to generate the final masks. Fifth, the detection head is used to predict the object bounding boxes. It takes the base feature pyramid and the instance-specific feature maps as input and generates the bounding box coordinates. Finally, the mask head takes the base feature pyramid and the instance-specific feature maps as input and generates the class-specific mask coefficients.

The Yolact model is trained using three different types of loss: classification loss, box regression loss, and mask loss. Both the classification loss and box regression loss are calculated in the same way as outlined in the reference \cite{liu2016ssd}. To compute the mask loss, the pixel-wise binary cross-entropy between the predicted masks and the ground truth masks was calculated \cite{bolya2019yolact}.

\subsection{Extraction of Geometric Properties}
As shown in Figure, the Yolact \cite{bolya2019yolact} is utilized to create a unique mask for each bacterial cell, which is then transformed into a binary image to extract the number of cells, area, length, and width. Moment Invariant \cite{mukundan1998moment} is a technique to extract the global features for form recognition and identification analysis. We employ the moment of invariant approach to extract the geometric properties of the individual bacterial cells from the binary image of segmented cells. A flowchart of this process is shown in Fig. \ref{fig_1}(d). Given a pixel intensity of an image array $f(x,y)$ with an image dimension of $w \times h$ pixels, moment invariants are defined by \cite{horn1986robot} \cite{rocha2002image}: 
\begin{equation}
m_{pq}=\sum_{x=1}^w\sum_{y=1}^h x^p y^q f(x, y),
\label{eq:eq1}
\end{equation}
The order of the moment is (p + q) where p and q are both natural numbers. In the case of the binary image, the pixel intensity is either 0 or 1. 

The area of a segmented cell in the binary image is given by the zeroth moment ($m_{00}$) of equation \ref{eq:eq1} is given by \cite{horn1986robot} \cite{teague1980image},
\begin{align}
m_{00}&=\sum_{x=1}^w\sum_{y=1}^h (x)^0(y)^0 f(x, y) \nonumber\\
         &=\sum_{x=1}^w\sum_{y=1}^h f(x, y),
\end{align}
where $m_{00}$ is the area of the object. The first-order and the second-order moments are described in
\begin{align}
m_{10}=\sum_{x=1}^w \sum_{y=1}^h xf(x, y), &&\text{ }m_{01}=\sum_{x=1}^w \sum_{y=1}^h yf(x, y), \nonumber
\end{align}
\begin{align}
m_{11}=\sum_{x=1}^w \sum_{y=1}^h xyf(x, y),&& m_{20}=\sum_{x=1}^w \sum_{y=1}^h x^2f(x, y), \nonumber
\end{align}
\begin{align}
m_{02}=\sum_{x=1}^w \sum_{y=1}^h y^2f(x, y), \nonumber
\end{align}
where $m_{10}$, $m_{01}$, and $m_{11}$ are the first order moment and $m_{20}$ and $m_{20}$ are the second order moment of equation \ref{eq:eq1}. 

The above equations are used to determine the length and width of each bacterial cell. However, the cells in a binary image have unusual shapes, making it challenging to determine the object's length and width. Moment invariants provide us the ability to construct the equivalent ellipse that fits these objects the best, as shown in Figure. We then calculate the semi-major axis ($a$) and semi-minor axis ($b$) of the ellipse by the following equations \cite{teague1980image} \cite{teague1980image},
\begin{align}
a=\sqrt{\frac{m_{20}+m_{02}+\sqrt{(m_{20}-m_{02})^2+4m_{11}^2}}{m_{00}/2}},
\end{align}
\begin{align}
b=\sqrt{\frac{m_{20}+m_{02}-\sqrt{(m_{20}-m_{02})^2+4m_{11}^2}}{m_{00}/2}},
\end{align}
The length and width of each cell are $2a$ and $2b$, respectively. We can calculate the perimeter of the object in the binary image using the following formula \cite{tomkys1902formula},
\begin{align}
2\pi\sqrt{\frac{a^2+b^2}{2}}.
\end{align}

\section{Results and Discussion}
\subsection{Key Findings}
Compared to our previous Mask-RCNN approach, the DLv3+ method, and manual measurement by domain experts, the Yolact-based model executes the geometric shape extraction of the bacterial cells 2.1x, 6.8x, and 446x faster. Furthermore, our numerical results show that the Yolact model greatly outperforms MRCNN and DLv3+ in measuring geometric features such as area, length, width, and the number of bacterial cells in biofilm microscope images.

\subsection{Data Generation and Collection}
The 2D-BEST center provides the biofilm microscopy image datasets (SEM images). On the surfaces of mild steel and copper materials, SEM imaging was used to characterize the biofilm and corrosion products of DA-G20. The following ingredients were used to establish DA-G20 axenic cultures: sodium lactate (6.8\%), dehydrated calcium chloride (0.06\%), sodium citrate (0.3), sodium sulfate (4.5), magnesium sulfate (2), ammonium chloride (1), potassium phosphate monobasic (0.5), and yeast extract (1). Sterile N2 gas was used to deoxygenate the sterile lactate media for 20 minutes at 15 psi and cultures were grown at 30 °C under shaking conditions for 48 hours at 125 rpm. Mild steel samples covered in the biofilm were subsequently soaked in 3 percent glutaraldehyde in cacodylate buffer (0.1 M, pH 7.2) for two hours. On samples of 56.2\% cold-worked copper that were exposed to corrosion cells, we grew DA-G20 cells. The samples were also cultured at the same time as the testing, which lasted 70 days. With sodium cacodylate buffer and distilled water, the treated samples were washed. Fig. \ref{fig_1}(a) shows the image generation process. The anaerobic chamber was used to grow biofilm on the mild steel and copper surfaces. After different days of exposure to the bacterial cells, the images were captured by a scanning electron microscope. We used 66 SEM images of the biofilm to train and test the deep-learning models, which are discussed next. The biofilm images have dimensions of $229$ pixels in height and $256$ pixels in width.

\subsection{Dataset Preprocessing}
On a variety of computer vision tasks \cite{baumgart1975polyhedron} \cite{shapiro2001computer} \cite{forsyth2002computer}, deep convolutional neural networks have exhibited astounding performance. However, these networks significantly rely on large datasets and the quality of the datasets. Our biofilm dataset only contains 66 SEM images; some have blurred borders between bacterial cells and the background, which may lead to the overfitting of the deep learning models. Overfitting \cite{dietterich1995overfitting} is a phenomenon that occurs when a network learns a function from smaller and poor-quality datasets that gives accurate predictions for the training data but not for the unseen data. Therefore, the volume and quality of the dataset determine how well a deep-learning neural network performs. Due to the small volume of the biofilm dataset and the "blurred" borders in some of the images (shown in Fig. \ref{fig_1}(a)), the deep learning model's ability to recognize and segment cells may suffer. In order to tackle these issues, we apply data augmentation techniques \cite{shorten2019survey}, a data-space solution to the problem of the small size of datasets, like rotation, mirror, vertical flipping, image cropping, and image scale, to increase the data volume and prevent the model from becoming overfit to the training dataset, as shown in Fig. \ref{fig_1}(c). In this study, image scale augmentation was used to randomly select a short image within a dimension range, crop patches from the original images randomly, mirror the image with a probability of 1/2, flip it vertically with a probability of 1/2, and rotate it 90 and 180 degrees.
\begin{table}[!h]
\caption{The training parameters used for Yolact.}
\centering
\begin{tabular}{l l} 
\toprule 
\textbf{Parameter} & \textbf{Value} \\ 
\midrule
    Initial learning rate & 0.0001\\
    Learning momentum & 0.9\\
    Weight decay & 0.0005\\
    Learning steps & 28000, 36000, 50000\\
    Min-batch size & 4\\
    Maximum iterations & 50000\\
    Iterations per training epoch  & 10\\
\bottomrule
\end{tabular}
\label{tab_dltune}
\end{table}
\begin{table}
\caption{Average Precision(AP in \%) on validation dataset with and without data augmentation during training.}
\centering
\begin{tabular}{c c c c c }
\toprule
\textbf{} & \textbf{} & \textbf{$AP$}  & \textbf{$AP_{50}$}  & \textbf{$AP_{75}$}\\
\midrule
Without & box & 7.23 & 16.61 & 9.72 \\
data augmentation & mask & 8.41 & 18.09 &  10.95 \\
\toprule
With & box & 26.78 & 58.35 &  32.61 \\
data augmentation & mask & 27.19 & 60.03 &  33.63 \\
\bottomrule
\end{tabular}
\label{tab1}
\end{table}

\begin{table}[!h]
\caption{Computing the number of bacterial cells without or with CLAHE using Yolact with ResNet-101. T1, T2, and T3 represent Test Image 1, Test Image 2, and Test Image 3, respectively. GT= Ground Truth} 
\centering
\begin{tabular}{l c c c} 
\toprule 
& \multicolumn{3}{c}{\textbf{Number of Cells}} \\
\cmidrule(l){2-4} 
\thead{Test Image} & \thead{GT} & \thead{Without CLAHE} & \thead{With CLAHE}\\ 
\midrule
T1 & 63 & 44 & 49 \\ 
T2 & 55 & 45 & 50 \\ 
T3 & 50 & 41 & 48 \\
\bottomrule
\end{tabular}
\label{tab2}
\end{table}
As well, the Contrast Limited Adaptive Histogram Equalization (CLAHE) algorithm \cite{pizer1987adaptive} is applied for contrast enhancement to detect more bacterial cells. Contrast Limited Adaptive Histogram Equalization (CLAHE) is based on adaptive histogram equalization, where two primary parameters: the block size (N) and clip limit (CL), are chosen by users to influence image quality. The flow of CLAHE is started by dividing an input image into small blocks, and the pixel's context of the small region is taken into account when calculating the histogram. Thus, the intensity of the pixel is changed to a value corresponding to its rank in the local intensity histogram and placed inside the display range to get enhanced images. 

\begin{table*}[!h]
\caption{Estimated average size characteristics of bacterial cells. T1, T2, T3, T4, T5, and T6 represent Test Image 1, Test Image 2, Test Image 3, Test Image 4, Test Image 5, and Test Image 6, respectively. GT= Ground Truth, DLv3+= DeepLabv3+, MCNN= MRCNN. (Best results are in bold. The results are presented in the format of $(mean \pm std)$)} 
\centering
\begin{tabular}{c c c c c c c}
\toprule 
\thead{Test Images} & \thead{Method} & \thead{No. \\of Cells}  & \thead{Avg. \\Area \\($\mu m^2$)} & \thead{Avg. \\Perimeter \\($\mu m$)} & \thead{Avg. \\Length \\($\mu m$)} & \thead{Avg. \\Width \\($\mu m$)}\\ 
\midrule 
\multicolumn{7}{c}{\textbf{%
 Results from DA-G20 biofilm images developed on mild steel surfaces}} \\
\toprule
\multirow{4}{1cm}{T1} & GT & 63 & 0.89 $\pm$ 0.30 & 4.32 $\pm$ 1.05 & 1.62 $\pm$ 0.45 & 0.69 $\pm$ 0.13 \\
& DLv3+ & 48 & 0.63 $\pm$ 0.21 & 3.35 $\pm$ 0.76 & 1.40 $\pm$ 0.41 & 0.64 $\pm$ 0.11\\
& MRCNN & \textbf{49} & 0.95 $\pm$ 0.28 & \textbf{4.31 $\pm$ 1.02} & 1.56 $\pm$ 0.42 & 0.75 $\pm$ 0.14\\
& Yolact & \textbf{49} & \textbf{0.91 $\pm$ 0.31} & 4.48 $\pm$ 1.05 & \textbf{1.65 $\pm$ 0.44} & \textbf{0.73 $\pm$ 0.14} \\
\midrule
\multirow{4}{1cm}{T2} & GT & 55 & 1.01 $\pm$ 0.40 & 4.57 $\pm$ 1.34 & 1.70 $\pm$ 0.48 & 0.74 $\pm$ 0.16 \\
& DLv3+ & 45 & 0.74 $\pm$ 0.23 & 3.64 $\pm$ 0.78 & 1.53 $\pm$ 0.40 & 0.69 $\pm$ 0.10\\ 
& MRCNN & \textbf{52} & 1.15 $\pm$ 0.52  & 4.23 $\pm$ 1.39 & 1.62 $\pm$ 0.64 & \textbf{0.76 $\pm$ 0.22} \\
& Yolact & 50 & \textbf{1.07 $\pm$ 0.42} & \textbf{4.63 $\pm$ 1.35} & \textbf{1.76 $\pm$ 0.50} & 0.79 $\pm$ 0.16 \\
\midrule
\multirow{4}{1cm}{T3} & GT & 50 & 1.15 $\pm$ 0.41 & 4.91 $\pm$ 1.27 & 1.81 $\pm$ 0.45 & 0.80 $\pm$ 0.13 \\
& DLv3+ & 34 & 0.64 $\pm$ 0.32 & 3.11 $\pm$ 1.01 & 1.28 $\pm$ 0.51 & 0.67 $\pm$ 0.10\\ 
& MRCNN & 57 & \textbf{1.10 $\pm$ 0.40} & 4.24 $\pm$ 1.14 & 1.52 $\pm$ 0.50 & \textbf{0.81 $\pm$ 0.20} \\ 
& Yolact & \textbf{48} & 1.28 $\pm$ 0.42 & \textbf{4.94 $\pm$ 1.29} & \textbf{1.82 $\pm$ 0.44} & 0.92 $\pm$ 0.13\\
\midrule
\noalign{\vspace{0.5ex}}
\multicolumn{7}{c}{\textbf{%
 Results from DA-G20 biofilm images developed on copper surfaces}} \\
\midrule
\multirow{3}{1cm}{T4} & GT & 44 & 0.74 $\pm$ 0.31 & 3.91 $\pm$ 1.15 & 1.76 $\pm$ 0.57 & 0.59 $\pm$ 0.14 \\
& MRCNN & \textbf{42} & 0.80 $\pm$ 0.35 & \textbf{3.93 $\pm$ 1.22} & 1.49 $\pm$ 0.48 & 0.67 $\pm$ 0.24 \\ 
& Yolact & 48 & \textbf{0.70 $\pm$ 0.33} & 4.01 $\pm$ 1.26 & \textbf{1.56 $\pm$ 0.52} & \textbf{0.61 $\pm$ 0.22} \\
\midrule
\multirow{3}{1cm}{T5} & GT & 55 & 0.55 $\pm$ 0.28 & 3.27 $\pm$ 1.11 & 1.46 $\pm$ 0.56 & 0.53 $\pm$ 0.12 \\
& MRCNN & \textbf{53} & 0.63 $\pm$ 0.41 & 3.51 $\pm$ 1.46 & 1.28 $\pm$ 0.42 & 0.63 $\pm$ 0.27 \\ 
& Yolact & 49 & \textbf{0.56 $\pm$ 0.31}  & \textbf{3.50 $\pm$ 1.27} & \textbf{1.36 $\pm$ 0.51} & \textbf{0.56 $\pm$ 0.20} \\
\midrule
\multirow{3}{1cm}{T6} & GT & 64 & 0.76 $\pm$ 0.33 & 4.23 $\pm$ 1.53 & 1.94 $\pm$ 0.79 & 0.59 $\pm$ 0.14 \\
& MRCNN & 58 & \textbf{0.72 $\pm$ 0.23} & 3.67 $\pm$ 1.08 & 1.41 $\pm$ 0.50 & 0.62 $\pm$ 0.17 \\ 
& Yolact & \textbf{61} & 0.62 $\pm$ 0.30 & \textbf{3.71 $\pm$ 1.28} & \textbf{1.45 $\pm$ 0.51} & \textbf{0.57 $\pm$ 0.16} \\
\bottomrule 
\end{tabular}
\label{tab4}
\end{table*}

\begin{table*}
\caption{Segmentation efficiency ($F_1$ Score in \%) evaluation performance. T1, T2, T3, T4, T5, and T6 represent Test Image 1, Test Image 2, Test Image 3, Test Image 4, Test Image 5, and Test Image 6, respectively. DLv3+= DeepLabv3+, MCNN= MRCNN. (Best results are in bold. The results are presented in the format of $(mean \pm std)$)}
\centering
\begin{tabular}{c c c c c c c c c}
\toprule
\textbf{Method} & \textbf{Backbone} & \textbf{T1}  & \textbf{T2}  & \textbf{T3} & \textbf{T4} & \textbf{T5} & \textbf{T6} & \textbf{Overall}   \\
\midrule
DLv3+ & ResNet-50 & 76.29 $\pm$ 0.56 & 80.84 $\pm$ 0.24 &  67.89 $\pm$ 0.29 & 76.19 $\pm$ 0.41 & 77.33 $\pm$ 0.19 & 72.56 $\pm$ 0.36 & 75.18 $\pm$ 0.34\\
MRCNN & ResNet-101 & 74.97 $\pm$ 0.36 & 76.37 $\pm$ 0.49 & 80.33 $\pm$ 0.11 & 82.63 $\pm$ 0.91 & 75.57 $\pm$ 0.75 & 76.14 $\pm$ 0.31 & 77.67 $\pm$ 0.49\\
\toprule
Yolact & Darknet-53 & 83.15 $\pm$ 0.26 & 85.74 $\pm$ 0.03 & \textbf{85.02 $\pm$ 0.13} & 88.37 $\pm$ 0.21 & \textbf{83.94 $\pm$ 0.09} & 80.41 $\pm$ 0.35 & 84.44 $\pm$ 0.18\\
 & ResNet-50 & 81.83 $\pm$ 0.17 & \textbf{86.97 $\pm$ 0.27} &  84.36 $\pm$ 0.15 & 89.56 $\pm$ 0.20 & 82.63 $\pm$ 0.22 & 81.69 $\pm$ 0.03 & 84.50 $\pm$ 0.17 \\
 & \textbf{ResNet-101} & \textbf{83.88 $\pm$ 0.13} & 86.83 $\pm$ 0.35 &  84.52 $\pm$ 0.18 & \textbf{90.52 $\pm$ 0.09} & 83.67 $\pm$ 0.11 & \textbf{82.23 $\pm$ 0.23} & \textbf{85.28 $\pm$ 0.18}\\ 
\bottomrule
\end{tabular}
\label{tab5}
\end{table*}

\begin{table*}
\caption{The execution time of each method for extracting geometric properties from the test images where m is minute and s is seconds. T1, T2, T3, T4, T5, and T6 represent Test Image 1, Test Image 2, Test Image 3, Test Image 4, Test Image 5, and Test Image 6, respectively. DLv3+= DeepLabv3+, MCNN= MRCNN. (Best results are in bold. The results are presented in the format of $(mean \pm std)$)}
\centering
\begin{tabular}{@{}p{1.2cm}@{} @{}p{1.6cm}@{} @{}p{2.1cm}@{} @{}p{2.1cm}@{} @{}p{2.1cm}@{} @{}p{2.1cm}@{} @{}p{2.1cm}@{} @{}p{2.1cm}@{} @{}p{2.1cm}@{}}
\toprule
\textbf{Method}  & \textbf{Backbone} & \textbf{T1}  & \textbf{T2}  & \textbf{T3} & \textbf{T4} & \textbf{T5} & \textbf{T6} & \textbf{Overall} \\
\midrule
Manual & - - - & 23m 34s $\pm$ 4m & 20m 15s $\pm$ 3m &  18m 54s $\pm$ 2m & 16m 30s $\pm$ 3m & 19m 49s $\pm$ 2m & 22m 11s $\pm$ 4m & 20m 12s $\pm$ 3m\\
DLv3+ & ResNet-50 & (18.97 $\pm$ 1.51)s & (17.87 $\pm$ 2.01)s &  (17.08 $\pm$ 1.89)s & (18.33 $\pm$ 1.44)s & (18.89 $\pm$ 1.96)s & (19.61 $\pm$ 2.36)s & (18.46 $\pm$ 1.86)s\\
MRCNN & ResNet-101 & (5.68 $\pm$ 1.11)s & (5.50 $\pm$ 1.57)s &  (5.48 $\pm$ 1.01)s & (5.07 $\pm$ 1.27)s & (5.67 $\pm$ 1.31)s & (6.02 $\pm$ 1.29)s & (5.56 $\pm$ 1.26)s\\
\toprule
Yolact & Darknet-53 & (2.91 $\pm$ 1.26)s & (2.83 $\pm$ 0.03)s &  (2.81 $\pm$ 0.13)s & \textbf{(2.69 $\pm$ 1.77)s} & (3.05 $\pm$ 1.38)s & (2.91 $\pm$ 1.84)s & (2.87 $\pm$ 1.07)s\\
 & ResNet-50 & (3.01 $\pm$ 1.77)s & (2.77 $\pm$ 1.23)s &  \textbf{(2.37 $\pm$ 1.09)s} & (2.81 $\pm$ 1.56)s & (3.31 $\pm$ 1.07)s & \textbf{(2.39 $\pm$ 1.47)s} & (2.76 $\pm$ 1.37)s \\
 & \textbf{ResNet-101} & \textbf{(2.56 $\pm$ 0.96)s} & \textbf{(2.67 $\pm$ 1.03)s} &  (2.59 $\pm$ 1.26)s & (2.71 $\pm$ 1.15)s & \textbf{(2.89 $\pm$ 1.02)s} & (2.88 $\pm$ 1.55)s & \textbf{(2.72 $\pm$ 1.16)s}\\ 
\bottomrule
\end{tabular}
\label{tab6}
\end{table*}

\subsection{Bacterial Cell Segmentation via Yolact}
The Yolact model was trained on a total of 66 images. The DA-G20 biofilms that were formed on mild steel surfaces provided 45 training images, 15 validation images, and 3 test images referred to as Test Image 1 (T1), Test Image 2 (T2), and Test Image 3(T3). The 3 more test images (referred to as Test Image 4 (T4), Test Image 5 (T5), and Test Image 6 (T6)) were obtained from copper-surfaced DA-G20 biofilms. It should be noted that the training stage of the deep learning models does not use images from copper-surfaced DA-G20 biofilms in order to evaluate their robustness. With the help of subject matter experts at the 2D-BEST center, we label the bacterial cells and backdrop in the biofilm picture datasets using the COCO Annotator tool \cite{cocoannotator}. Then, we used that labeling dataset for training, validating, and testing the model. On the same machine that we used for MRCNN and DLv3+, we trained the Yolact model using an onboard GPU (NVIDIA GTX 16 Series, 6 GB memory), enabling GPU-based acceleration with CUDA to reduce the training period. During the training process, we chose different hyperparameter values to train the model. Hyperparameters are variables whose values influence the learning process and define the model parameter values such as the number of iterations, learning rate, batch size, etc that a learning algorithm ultimately learns. The training process didn't take much longer since we used a small dataset with pre-trained weights. We tuned the hyperparameters of the Yolact model via trial and error with different values along with the ResNet-50 \cite{he2016deep}, ResNet-101 \cite{he2016deep}, and DarkNet-53 \cite{redmon2018yolov3} backbone networks.  We got our best accuracy for the following values of the hyperparameter as shown in Table \ref{tab_dltune}. Every 1000 iterations, the model's weights were saved, and they were subsequently utilized to analyze the performance results on the test datasets. We trained the network with a batch size of $4$, a learning rate of $1 \times 10^{-3}$, and $50,000$ iterations. 

The loss function curves of the training process for bounding box, segmentation (mask loss), and classification are shown in Fig. \ref{fig_1}(e).  Our dataset was increased by using data augmentation methods, as previously discussed. The Yolact model was trained and evaluated without and with data augmentation methods. To quantitatively verify the impact of data augmentation on unseen data (validation data), average precision (AP) \cite{kishida2005property} is used to evaluate the model's efficiency with and without data augmentation. Average Precision (AP) of validation results on training dataset with and without data augmentation as shown in Table \ref{tab1}. From Table \ref{tab1}, the average precision of the segmented mask and the bounding box without data augmentation on validation data is 8.41\% and 7.23\%, respectively, whereas these values increase to 26.76\% and 27.19\% with data augmentation. This result suggests that the validation efficiencies are significantly increased with the data augmentation methods that overcome the model's overfitting issue. Next, we preprocessed the training dataset using the CLAHE approach to determine how CLAHE affected the Yolact network's ability to count bacterial cells. We trained the deep neural network (Yolact) both with and without CLAHE and the results are shown in Table\ref{tab2}. From Table\ref{tab2}, the number of bacterial cells with CLAHE is higher than without CLAHE. Subject matter experts at the 2D-BEST center tallied the number of bacterial cells in T1, T2, and T3 and found that there were 44, 45, and 41, respectively which we considered as ground truth. The Yolact model counts the number of bacterial cells in T1, T2, and T3 as 44, 45, and 41 without CLAHE, while with the CLAHE method, these numbers are 49, 50, and 48, respectively. The results with the CLAHE method are closer to ground truth. The ground truths and the cell segmentation results from the deep learning techniques are shown in Fig. \ref{img3} and Fig. \ref{fig_4}(a).

\subsection{Estimated Geometric Properties}
The measurement of bacterial cell sizes is important since the size of a bacterial cell varies depending on the growth conditions. However, their various phenotypes are caused by their diverse gene expression, which determines the mechanism of their different cell sizes. These ambiguous genotypical and phenotypic changes at the materials-microbe interface are reliant on the interaction of microbial biofilms. We use the moment of invariants method to determine the size parameters of the segmented bacterial cells, including their area, length, width, and perimeter. Three test images (T1, T2, T3) of DA-G20 biofilms developed on mild steel surfaces from related MIC studies were used to assess the effectiveness of all the techniques covered in this study. We additionally examine the model's effectiveness for cell segmentation and cell size estimate on the microbial corrosion system where DA-G20 cells are cultured on copper substrates. The results of the geometric feature extraction are presented in Table \ref{tab4}. The findings in Table \ref{tab4} show that Yolact's performance in estimating the size properties of bacterial cells on biofilm images of both mild steel surfaces and copper surfaces is close to ground truth. 

Fig. \ref{fig_4} is a visual representation of the average values and the corresponding error bars (standard deviation) for the estimated geometric properties of all the segmented cells from T1, T2, and T3. The error bars indicate the degree of uncertainty or variability in the data. The geometric properties depicted in the figure include the characteristics such as the number of bacterial cells, area, length, width, and perimeter of the segmented cells. Based on Figures \ref{fig_4}(b), \ref{fig_4}(c), \ref{fig_4}(d), and \ref{fig_4}(e), it appears that the MRCNN and DLv3+ models have a large degree of variability in their estimates of the area, length, width, and perimeter of segmented bacterial cells, which is not present in the actual measurements. This suggests that the MRCNN and DLv3+ models may not be as accurate as other methods such as Yolact. However, when counting the number of cells, MRCNN and Yolact have similar performance levels compared to the ground truth measurements while results from the DLv3+ technique are unsatisfactory as shown in Figure \ref{fig_4}(a). Overall, the Yolact method appears to be the most accurate for estimating both the geometric properties of cells and counting their number as shown in Fig. \ref{fig_4}. As discussed earlier, DA-G20 cells grown on copper surfaces were used for cell segmentation and size estimation to assess the robustness of our proposed model. The raw SEM images (T4, T5, and T6) from copper surfaces, cell segmentation results from the MRCNN and Yolact approaches, and the cell size estimation results are shown in Table \ref{tab4} and Figures \ref{fig_4}(b), \ref{fig_4}(c), and \ref{fig_4}(d). It is important to point out that we have only evaluated the MRCNN model with Yolact for test images 4, 5, and 6, as MRCNN demonstrated the best performance compared to DLv3+. The outcomes in Figures \ref{fig_4}(b), \ref{fig_4}(c), and \ref{fig_4}(d) show that Yolact works rather well when estimating the size properties of bacterial cells compared to the results of MRCNN. In conclusion, Fig. \ref{fig_4} and Table \ref{tab4}'s data show that Yolact is effective at segmenting and measuring the size of cells in biofilms formed on diverse metal surfaces. These various metal systems provide proof-of-concept that the Yolact method can be used to analyze the structural relationships in different bacterial systems by displaying the adaptability of this method.

\subsection{Model Performance}
It is essential to determine the level of trustworthiness of a trained model when making predictions on unseen data. In this case, we apply a cross-validation technique known as the Dice similarity coefficient ($F_1-Score$) to evaluate the model's segmentation accuracy of bacterial cells on the test images. As well as, we also evaluate the time taken for cell segmentation and extraction of size properties for each method discussed, including manually, DLv3+, MRCNN, and Yolact.

The $F_1 -score$ \cite{chinchor1993muc} is a measure of a model's accuracy that balances precision and recall. It is a commonly used metric in the field of machine learning and is particularly useful for binary classification problems. The $F_1 -score$ is the harmonic mean of precision and recall, where the best score is 100\% and the worst is 0\%. Precision is the proportion of true positive predictions (i.e. the number of times the model correctly predicted a positive outcome) out of all positive predictions made by the model. The recall is the proportion of true positive predictions out of all actual positive outcomes. A high $F_1 -score$ indicates that the model has high accuracy and is able to balance precision and recall well. In a binary classification problem, it means that the model is able to correctly identify the positive instances (True positive) and also able to minimize the false positives. We evaluate the $F_1 -score$ \cite{chinchor1993muc} using the following equation:
\begin{align}
F_1 -score = \frac{2 \times precision \times recall}{precision + recall},
\end{align}
where $prcision$ and $recall$ are calculated by the following equations:
\begin{align}
precision = \frac{TP}{TP + FP},
\end{align}

\begin{align}
recall = \frac{TP}{TP + FN},
\end{align}
where True Positive ($TP$) is the number of bacterial cells that were correctly identified by the model, a False Negative ($FN$) is the number of bacterial cells that were not identified by the model, and a False Positive ($FP$) is the number of cells that were incorrectly identified by the model. These values of $TP$, $FN$, and $FP$ are calculated by using Python programming and are determined by averaging the results over each test image. The results of each method applied to T1, T2, T3, T4, T5, and T6 are presented in Table \ref{tab5}. These results indicate that Yolact has a better performance than the other methods considered in this study. The $F_1-Score$ of Yolact with ResNet-101 backbone network achieves $85.28\%$ where DLv3+ with ResNet-50 and MRCNN with ResNet-101 are $75.56\%$ and $77.67\%$, respectively.

Next, we measure the execution times for the manual measurement method used by domain experts at the 2D-BEST center, as well as the geometric properties measured by them. On the same system, we utilize the 'timeit' library for MRCNN and Yolact and the 'timeElapsed' function for DeepLabv3+ to calculate the execution timings.  The results are summarized in Table \ref{tab6}, which shows the average time taken to complete the experiment for all cells in each of the six test images. According to the results, Yolact models outperform the manual method, DLv3+, and MRCNN not only for D20 biofilm images on mild steel surfaces but also for D20 biofilm images on copper surfaces. Particularly, the Yolact model outperforms the MRCNN, DLv3+, and manual process by $2.1\times$, $6.8\times$, and $246\times$, respectively. In conclusion, the Yolact approach is the best choice (among the methods considered here) both in terms of segmentation accuracy and execution time.

\section{Conclusion}
In summary, we developed the BioflimScanner tool which is a deep learning-based image segmentation approach using the Yolact architectures with the moment invariants method to automate the extraction of geometric size properties of bacterial cells in biofilms. To automate the process of evaluating structural changes in the biofilms in response to the coatings, high-throughput microscopy image characterization methods are needed for fast and efficient screening and selection of protective coatings against the MIC effects of biofilms. To help automate the process of measuring structural changes in biofilms, we used a neural network architecture (Yolact) with moment invariants, BiofilmScanner, to segment bacterial cells and extract geometric properties of segmented cells in microscopy images of biofilms. The study showed that BiofilmScanner outperforms both DeepLabV3+ and Mask R-CNN in terms of estimation accuracy and faster. More particularly, the F1-Score of the BiofilmScanner using the ResNet-101 backbone network was found to be 85.28\%, which is higher than the scores of the DeepLabv3+ model with ResNet-50 and the Mask R-CNN model with ResNet-101, which were 75.56\% and 77.67\% respectively. Additionally, the BiofilmScanner tool is 2.1x, 6.8x, and 246x, respectively, faster than our earlier Mask-RCNN, DLv3+, and manual measurement by the domain experts.  Our research has shown that the methods we developed for segmenting and measuring bacterial cells in biofilms can be applied to other types of biofilms as well. Specifically, we have demonstrated that, without any additional training, our methods can be used to analyze biofilms of other types of bacteria, such as \textit{E. Coli}, \textit{P. aeruginosa}, and \textit{B. subtilis}, as long as the shape and structure of the bacterial cells in those biofilms are similar to that of the DA-G20 biofilm that we used in our study.


%



\ifCLASSOPTIONcompsoc
  \section*{Acknowledgments}
\else
  \section*{Acknowledgment}
\fi

The authors acknowledge funding support from NSF RII T-2 FEC award \#1920954. S. Ragi would like to acknowledge NSF RII T-1 FEC award \#1849206 for a seed grant that partially supported this study. V. Gadhamshetty would like to acknowledge partial support from NSF CAREER award \#1454102. Dr. Gadhamshetty’s group is thankful to Dr. Bharat Jasthi, Materials and Metallurgical Engineering (MET), SD Mines for providing copper samples for the dislocation experiments. 

\section*{Data Availability}
The data that support the findings of this study are available from the corresponding author upon request. 

\section*{Code Availability}
The source code, the trained network weights, and the training data are available at \url{https://github.com/hafizur-r/BiofilmScanner-v0.2} 

\section*{CORRESPONDING AUTHOR}
Md Hafizur Rahman – Embedded Engineer, GM Global Technical Center, Cole Engineering Center Tower, 29755 Louis Chevrolet Road, Warren, MI 48093; Phone: (605) 391-0506; Email: \href{mailto:hafizur.raj@gmail.com}{hafizur.raj@gmail.com} or \href{mailto:mdhafizur.rahman@gm.com}{mdhafizur.rahman@gm.com}

\ifCLASSOPTIONcaptionsoff
  \newpage
\fi



%

\bibliography{refs.bib}
\bibliographystyle{IEEEtran}

%
\begin{IEEEbiography}[{\includegraphics[width=1in,height=1.25in,clip,keepaspectratio]{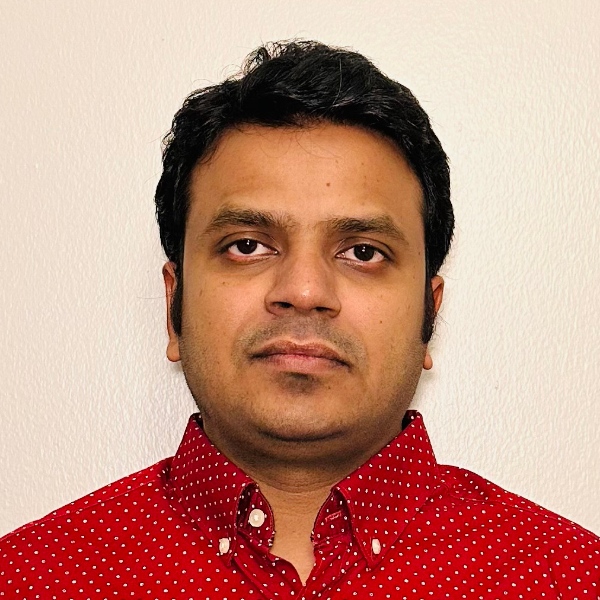}}]{Md Hafizuar Rahman}
has completed his M.S. in Electrical Engineering at South Dakota Mines and his B.Sc. in Electrical and Electronic Engineering at Pabna University of Science and Technology, Bangladesh. During his M.S., he worked on data-driven models for biofilm phenotype prediction on metal surfaces modified with 2D coatings. Currently, he is working as an Embedded Engineer (Contractual) at General Motors. His current research interests include deep learning, DL model quantization, and computer vision for autonomous vehicles.
\end{IEEEbiography}
\begin{IEEEbiography}[{\includegraphics[width=1in,height=1.25in,clip,keepaspectratio]{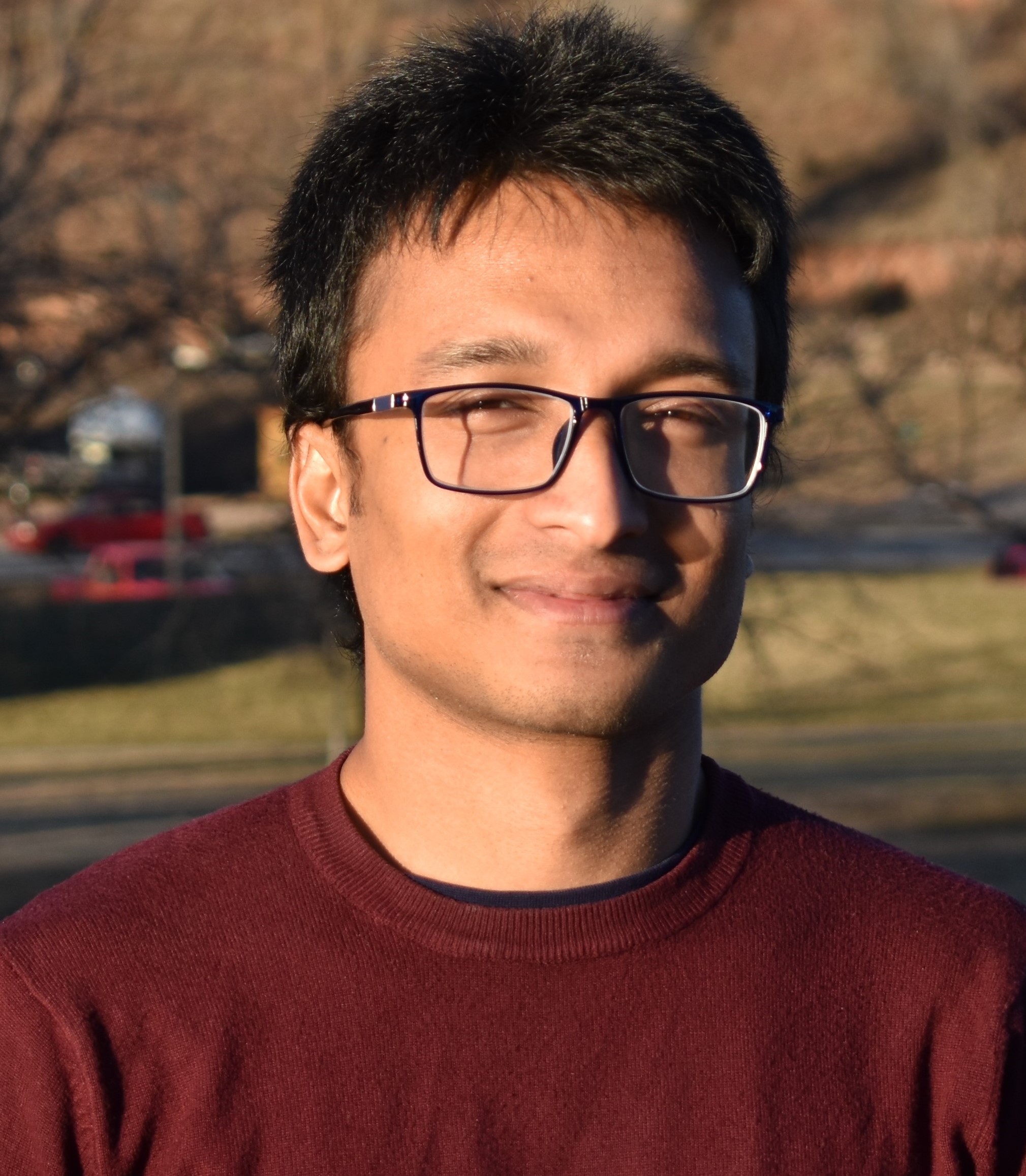}}]{Md Ali Azam} (Student Member, IEEE) was born in Tarabaria, Chartarapur, Pabna, Bangladesh, in 1992. He received the B.Sc. degree in electronics and telecommunication engineering from the Rajshahi University of Engineering and Technology, Bangladesh, in 2014, and the M.S. degree in electrical engineering from the South Dakota School of Mines and Technology, Rapid City, SD, USA, in 2020. From 2018 to 2020, he was a Graduate Assistant at SDSMT, where he worked as a Graduate Teaching Assistant and a Graduate Research Assistant during his M.S. studies. He worked as a System Engineer at a Telecommunication Company in Bangladesh, before attending SDSMT. During his M.S. studies, he published several papers.

\end{IEEEbiography}

\begin{IEEEbiography}[{\includegraphics[width=1in,height=1.25in,clip,keepaspectratio]{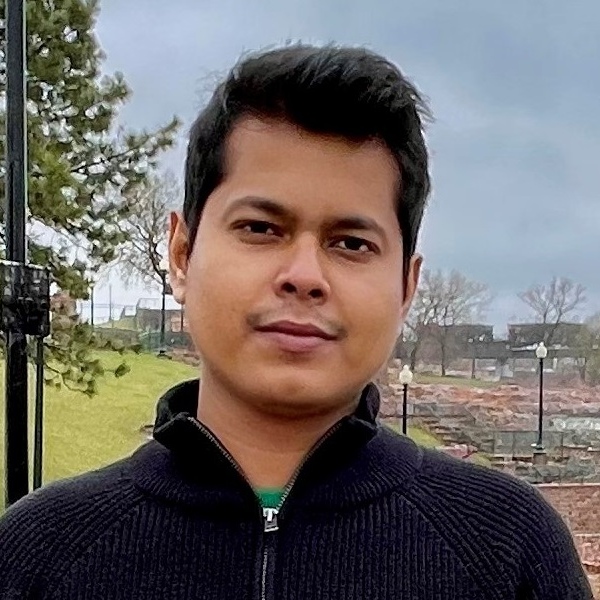}}]{Md Abir Hossen}
received the B.S. degree in Electrical and Electronics Engineering from American International University-Bangladesh, Dhaka, Bangladesh, in 2017 
and the M.S. degree in Electrical Engineering from South Dakota School of Mines and Technology, Rapid City, SD, USA in 2021. During his B.S. studies, he developed delivery robots with autonomous navigation capability to be deployed in hospitals on an a2i (a Bangladesh government program run from the Prime Minister’s office supported by UNDP and USAID) funded project. He was a graduate research assistant at the Unmanned and Swarm System (USS) laboratory during his M.S. studies from 2019-2021, where he developed AI-driven UAV-based multispectral sensing solution for agricultural soil monitoring. He is currently pursuing the Ph.D. degree in Computer Science at University of South Carolina, SC, USA. He is also working as a graduate research assistant at Artificial Intelligence and Systems Laboratory (AISys) and conducting research on finding root causes of functional faults in highly-configurable robotic systems through the lens of causality, and optimizing DNNs for the Europa Space Lander in collaboration with NASA. His research interest includes Autonomous and adaptive systems and Machine learning systems. 
\end{IEEEbiography}
\begin{IEEEbiography}[{\includegraphics[width=1in,height=1.25in,clip,keepaspectratio]{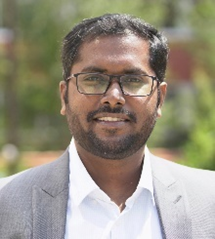}}]{Shankarachary Ragi}
was an assistant professor in the Electrical Engineering Department at South Dakota Mines, USA. He earned his Ph.D. degree in Electrical and Computer Engineering at Colorado State University, USA in 2014, and his B.Tech. and M.Tech. degrees in Electrical Engineering at the Indian Institute of Technology Madras, India in 2009. Before joining South Dakota Mines, Ragi has worked as a postdoctoral researcher in the mathematics department at Arizona State University, and prior to that, he worked as a Senior Controls Engineer at Cummins Emission Solutions. He is currently serving as senior personnel at the 2D-Materials for Biofilm Engineering, Science and Technology (2D-BEST) center funded by the National Science Foundation. His current research interests include machine learning, image analysis, robotics, and optimal control. Ragi has served as an Associate Editor for IEEE Access during 2017-2020. He has authored or co-authored over 26 peer-reviewed publications in various journals and conference proceedings. He is a senior member of the IEEE.      
\end{IEEEbiography}

\begin{IEEEbiography}[{\includegraphics[width=1 in,height=3in,clip,keepaspectratio]{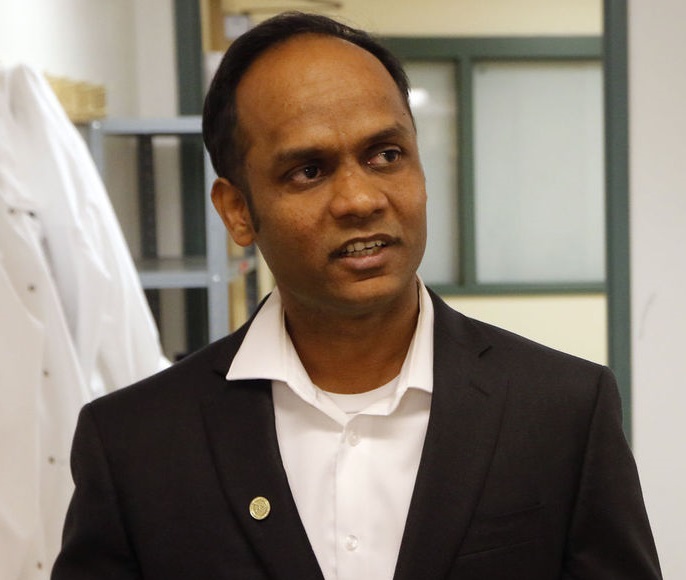}}]{Venkataramana Gadhamshetty}
 earned a Ph.D. degree in Civil and Environmental Engineering from New Mexico State University, an MS degree in Environmental Engineering from the National University of Singapore, and a BS degree in Chemical Technology from Osmania University. He is currently a Professor in the Civil and Environmental Engineering department at South Dakota Mines, USA. He has over a decade of teaching and research experience from South Dakota Mines, Rensselaer Polytechnic Institute, Florida Gulf Coast University, Air Force Research Laboratory, and industrial experience from Dupont Singapore Pte Ltd. He is a Board-Certified Environmental Engineer, a licensed Professional Engineer, and the chair of the ASCE EWRI Water Pollution Engineering Committee. He is a recipient of the National Science Foundation CAREER award (2015), South Dakota Mines Research Award (2016), and an invited Tedxtalk speaker for Rapid City in 2017. His research on bioelectrochemistry was featured by BBC, CNN, American Chemical Society, History Now, and 350 other large media outlets. He is a thrust area lead and core investigator (co-I) at the 2D-Materials for Biofilm Engineering, Science and Technology (2D-BEST) center and for other projects funded by NSF, NASA EPSCoR, and Electric Power Research Institute. He has served as an investigator or senior personnel for projects worth ~\$32 MM. His ongoing projects interrogate the fundamental phenomena at the interface of 2D materials and biofilms. Examples of practical outcomes from these projects include NASA microbial fuel cells and infinitesimally thin coatings for corrosion applications. 
\end{IEEEbiography}




\end{document}